\documentclass[10pt, a4paper]{article}
\usepackage[final]{lrec2026}
\usepackage{amsmath,amssymb,amsfonts}
\usepackage{algorithmic}
\usepackage{graphicx}
\usepackage{textcomp}
\usepackage{xcolor}
\usepackage{booktabs}
\usepackage{multirow}
\usepackage{array}
\usepackage{makecell}
\usepackage{subcaption}
\usepackage{url}
\usepackage{comment}
\usepackage{tabularx}
\usepackage{pifont}
\usepackage{enumitem}
\setlist[itemize]{nosep,leftmargin=*,topsep=2pt}
\setlist[enumerate]{nosep,leftmargin=*,topsep=2pt}

\title{QU-NLP at ArchEHR-QA 2026: Two-Stage QLoRA Fine-Tuning of Qwen3-4B for Patient-Oriented Clinical Question Answering and Evidence Sentence Alignment}

\name{Mohammad AL-Smadi}

\address{Qatar University \\
         Doha, Qatar \\
         malsmadi@qu.edu.qa}

\abstract{
We present a unified system addressing both Subtask~3 (answer generation)
and Subtask~4 (evidence sentence alignment) of the ArchEHR-QA Shared Task.
For Subtask~3, we apply two-stage Quantised Low-Rank Adaptation (QLoRA) to
Qwen3-4B loaded in 4-bit NF4 quantisation: first on 30{,}000
samples from the emrQA-MedSQuAD corpus to establish clinical domain competence,
then on the 20~annotated development cases to learn the task-specific output style.
Our system achieves an overall score of 32.87 on the official
test-2026 split (BLEU\,=\,9.42, ROUGE-L\,=\,27.04, SARI\,=\,55.42,
BERTScore\,=\,43.00, AlignScore\,=\,25.28, MEDCON\,=\,37.04).
For Subtask~4, we develop a weighted ensemble of three retrieval methods---
BM25 with relative thresholding, TF-IDF cosine similarity, and a fine-tuned
cross-encoder---to identify note sentences supporting a given gold answer,
achieving a micro-F1 of 67.16 on the 100-case test set.
Experiments reveal that both subtasks expose the same fundamental challenge: 20~annotated training
cases are insufficient to distinguish relevant from irrelevant clinical
sentences, pointing to data augmentation as the highest-leverage future direction.
}

\begin{document}

\sloppy

\maketitleabstract

%% ===========================================================
\section{Introduction}

Electronic Health Records (EHRs) contain the authoritative account of a
patient's hospitalisation: diagnosis, treatment decisions, test results,
and follow-up plans.
Yet this information remains largely inaccessible to patients and
caregivers because it is written in specialised medical language and
embedded within lengthy structured documents \citep{Johnson2016}.
Automatically generating plain-language answers to patient questions
from these records---\emph{patient-oriented QA}---is a task with
immediate clinical value \citep{Lin2003,Jeon2006}.
%\subsection{Clinical Question Answering from EHRs}

Question answering (QA) over clinical text has been studied since the creation of EHR-scale corpora. \citet{Pampari2018} introduced emrQA, the first large-scale QA benchmark built from MIMIC-III clinical notes, enabling span-extraction models
trained on structured EHR annotations.
\citet{Gu2021} showed that domain-adaptive pre-training on biomedical and
clinical text substantially improves downstream clinical NLP performance.
The shift from extractive to \emph{abstractive} generation changes
the evaluation paradigm: answers need not appear verbatim in the source
and may require simplification, synthesis, and register adaptation.
\citet{Lin2003} identified context and completeness as primary
determinants of answer quality, while \citet{Jeon2006} demonstrated
that non-textual structural signals predict answer reliability---both
insights motivating ensemble approaches that combine multiple retrieval
signals.
Ambient clinical intelligence datasets \citep{Yim2023} have further
expanded the scope to include visit note generation and evidence grounding.

Applying large language models (LLMs) to QA over clinical text raises specific
challenges.
Clinical notes are long and structurally heterogeneous, mixing narrative
prose with medication lists, laboratory values, and templated discharge
instructions \citep{Yim2023}.
Relevant information may be buried in a 60-sentence excerpt, and even a
correctly identified fact may be expressed in terminology the model has
not been optimised to reproduce.
Furthermore, hallucination---generating plausible but unsupported
clinical claims---is especially harmful in a medical context.

Parameter-efficient fine-tuning via LoRA \citep{Hu2022} and its
quantised variant QLoRA \citep{Dettmers2023} have made it feasible to
adapt billion-parameter models on a single consumer GPU. We contribute a two-stage QLoRA pipeline for patient-oriented clinical
question answering and a lexical--neural ensemble for evidence sentence
alignment, evaluated on the ArchEHR-QA Shared Task.
For answer generation (Subtask~3), we demonstrate that domain
pre-training on emrQA-MedSQuAD followed by task fine-tuning yields competitive performance using only a 4-bit quantised 4B-parameter model.
For evidence alignment (Subtask~4), we show that combining BM25,
TF-IDF, and a fine-tuned cross-encoder via weighted ensemble voting
achieves good results despite the small size of the training data of only 20 cases.

The remainder of this paper is organized as follows.
Section~\ref{sec:gap} identifies the research gaps motivating our approach.
Section~\ref{sec:task} explains the ArchEHR-QA Shared Task.
Section~\ref{sec:data} describes the datasets.
Section~\ref{sec:system} details the system architecture.
Section~\ref{sec:experiments} reports the achieved results.
Section~\ref{sec:conclusion} concludes this research.
 
%% ===========================================================

\section{Research Gap}
\label{sec:gap}
 
Despite growing interest in patient-oriented clinical QA, several gaps remain underexplored:

First, while some prior systems have used open-weight
models~\citep{soni-etal-2025-overview}, the top-ranked systems in the 2025 shared task often relied on proprietary LLMs combined with few-shot learning or synthetically generated examples~\citep{yoadsanit-etal-2025-lamar}, leaving open the question of how far open-weight fine-tuned models can perform under the same
annotation constraints. Second, the scarcity of annotated training data—only 20 development cases are provided in The ArchEHR-QA 2025 and 2026 shared task, with limited total labeled data overall—has not been systematically addressed. 
Third, To the best of our knowledge, the role of clinical domain adaptation as a substitute for task-specific annotation has not been systematically
evaluated in this task setting. Fourth, evidence grounding is often treated as an auxiliary component embedded within generated answers rather than as a standalone retrieval problem, making it difficult to isolate and improve grounding and generation independently.

In this research, we address these gaps by:
(i) fine-tuning an open-weight 4B model using QLoRA to eliminate dependence on proprietary APIs; (ii) employing a two-stage training strategy with domain adaptation
on emrQA-MedSQuAD~\citep{Pampari2018} followed by task-specific fine-tuning, providing a transparent and reproducible alternative to synthetic data generation; and (iii) designing a lexical--neural ensemble that combines BM25, TF-IDF, and a fine-tuned cross-encoder for Subtask~4, treating evidence
alignment as a first-class retrieval problem and enabling independent evaluation of each component.

%% ===========================================================

%% ===========================================================
\section{Task}
\label{sec:task}

The ArchEHR-QA Shared Task \citep{soni-etal-2026-archehr-qa} formalises this challenge
across two complementary subtasks.
\textbf{Subtask~3} requires producing a factually grounded answer of at
most 75~words in professional yet accessible language, given a
de-identified MIMIC-III~\citep{Johnson2016} and MIMIC-IV~\citep{johnson2023mimic} note excerpt ,
a patient question, and a clinician-interpreted reformulation.
\textbf{Subtask~4} requires identifying which note sentences constitute
the supporting evidence for a given gold answer, evaluated via micro-F1
over sentence-level binary citation decisions.
Together, the two subtasks model the full information access pipeline:
locate the evidence, then communicate it.

%\subsection{Parameter-Efficient Fine-Tuning}

\begin{comment} 

\subsection{Evidence Alignment and Retrieval}

Identifying note sentences that support a generated answer is related to
rationale extraction, reading-comprehension evidence selection, and
passage reranking.
Classical BM25 \citep{Robertson1994} and TF-IDF provide robust lexical
baselines whose performance is stable across document lengths, making
them reliable recall anchors.
Cross-encoders trained on MS-MARCO passage reranking offer high-precision
sentence scoring at inference time; fine-tuning them on domain-specific
evidence pairs further boosts precision on clinical text.
Ensemble methods combining lexical and neural signals are standard in
open-domain retrieval and transfer naturally to sentence-level citation
identification.
\end{comment}
%\subsection{Evaluation of Clinical Text Simplification}

Subtask~3 uses a group of evaluation metrics including: standard machine translation metrics BLEU \citep{Papineni2002} and ROUGE \citep{Lin2004}
measure n-gram overlap with reference answers but do not directly reward
simplification quality or clinical faithfulness, SARI \citep{Xu2016} specifically targets simplification by separately rewarding addition of plain-language terms, retention of source content, and deletion of unnecessary jargon, BERTScore \citep{Zhang2020} provides semantic similarity via contextual embeddings, capturing paraphrase beyond surface n-grams, AlignScore \citep{Zha2023} measures factual consistency between the generated answer and the source document, penalising hallucination, MEDCON assesses UMLS medical concept overlap between hypothesis and gold reference, rewarding precise clinical terminology. Subtask~4 is evaluated with standard information-retrieval metrics: micro-precision, micro-recall, and micro-F1 over sentence-level binary citation decisions.

%% ===========================================================

\begin{table}[t]
\centering
\small
\caption{ArchEHR-QA dataset statistics.}
\label{tab:dataset}
\setlength{\tabcolsep}{4pt}
\begin{tabular}{lrccc}
\toprule
\textbf{Split} & \textbf{Cases} & \textbf{IDs} \\
\midrule
dev         &  20 & 1--20    \\
test        & 100 & 21--120  \\
test-2026   &  47 & 121--167 \\
\midrule
\textbf{Total} & \textbf{167} & & &\\
\bottomrule
\end{tabular}
\end{table}
%% TABLE 2 — Subtask 3 example (Case 4, dev split, Cardiology)
%% Source: archehr-qa.xml + archehr-qa_key.json, case_id=4
\begin{table*}[t]
\centering
\small
\caption{Subtask~3 example from the \textbf{dev split} (Case~4, Cardiology,
MIMIC-III document 180932\_37135).
All 21 note sentences are shown verbatim from the XML excerpt
(\texttt{note\_excerpt\_sentences}); sentences truncated with \ldots{}
exceed 120 characters in the source.
Bold sentence numbers mark gold-cited sentences (Subtask~4 labels): \textbf{E}\,=\,essential,
\textbf{S}\,=\,supplementary.
MIMIC-III de-identified spans appear as \texttt{[**\ldots**]}.}
\label{tab:example_test}
\setlength{\tabcolsep}{4pt}
\renewcommand{\arraystretch}{1.15}
\begin{tabular}{@{}p{0.13\textwidth}p{0.82\textwidth}@{}}
\toprule
\textbf{Field} & \textbf{Case 4 — Cardiology}\\
\midrule
\textbf{Patient question} &
\textit{``My doctor performed a cardiac catherization.
Was this invasive, risky procedure necessary.''}\\
\addlinespace[2pt]
\textbf{Clinician question} &
\textit{``Why was cardiac catheterization recommended to the patient?''}\\
\addlinespace[3pt]
\textbf{Clinical note excerpt} &
1:\ History of Present Illness:\newline
2:\ On the cardiology service his abdominal pain, nausea, vomitting was felt to be secondary to congestive hepatopathy, his cough due to CHF vs asthma, and his syncope was felt to be secondary to a coughing spell\ldots\newline
3:\ His ICD interrogation was negative for any events.\newline
4:\ He was aggressively diuresed with a net 10 liters negative since admission.\newline
\textbf{[E]~5:}\ He underwent RHC for milrinone trial, which proved to be successful.\newline
6:\ His mean PCW went from 30 to 22, and his Fick C.I. went from 1.72 to 2.79.\newline
\textbf{[S]~7:}\ Brief Hospital Course: 48M with idiopathic dilated cardiomyopathy (EF 25\%, on coumadin \& s/p ICD placement), HTN, CKD (baseline Cr 1.3--1.6) \& asthma who presents with syncope, found to be in acute-on-chronic heart failure.\newline
8:\ \#) ACUTE-ON-CHRONIC SYSTOLIC HEART FAILURE\newline
\textbf{[S]~9:}\ Underlying exacerbation of chronic heart failure known idiopathic cardiomyopathy likely explanation for numerous presenting symptoms including syncope, abdominal pain, and dyspnea.\newline
\textbf{[E]~10:}\ Last echo showed LVEF = 25\%.\newline
\textbf{[E]~11:}\ He was admitted in the setting of low output heart failure, leading to increased intra-abdominal pressures causing congestive hepatopathy/abdominal pain/RUQ tenderness.\newline
12:\ Abdominal distension also thought to contribute to syncope.\newline
\textbf{[E]~13:}\ Cardiac output and wedge pressure significantly improved after milrinone infusion.\newline
14:\ In the ICU he was maintained on milrinone at a rate of 0.5 mcg/hr and transferred to the floor on this stable dose.\newline
15:\ Torsemide 80mg daily was restarted; he was discharged on milrinone and torsemide.\newline
16:\ Heart failure specialists had honest discussions with the patient about his long-term prognosis\ldots\newline
17:\ Discharge Instructions:\newline
\textbf{[E]~18:}\ You were admitted to the hospital with worsening heart failure.\newline
\textbf{[E]~19:}\ You had a cardiac catheterization that showed you would benefit from milrinone.\newline
\textbf{[E]~20:}\ You were started on a milrinone drip, with improvement in your heart's pump function.\newline
21:\ You were also diuresed for 6 liters of fluid.\\
\addlinespace[3pt]
\textbf{Gold answer} &
\textit{``The patient was recommended a cardiac catheterization for worsening heart failure confirmed by left ventricle ejection fraction of 25\% on his echocardiogram~[\textbf{5,10,18}]. He had low output heart failure which caused increasing intra-abdominal pressure resulting in congestive hepatopathy, abdominal pain, and right upper quadrant abdominal tenderness~[\textbf{11}]. The cardiac catheterization showed the patient needed milrinone for treatment~[\textbf{19}]. Milrinone infusion improved the patient's heart pump function by significantly improving cardiac output and wedge pressure~[\textbf{13,20}].''}\\
\bottomrule
\end{tabular}
\end{table*}

%% TABLE 3 — Subtask 4 example (Case 20, dev split, Neurology)
%% Source: archehr-qa.xml + archehr-qa_key.json, case_id=20
%% All 9 sentences fit in full; full citation trail shown per answer sentence
\begin{table*}[t]
\centering
\small
\caption{Subtask~4 example from the \textbf{dev split} (Case~20, Neurology,
MIMIC-IV document 22494097).The gold answer is split into its constituent sentences; each is annotated
with the note sentences it cites. Sentence-level relevance labels (\texttt{answers} field) are shown in
the rightmost column: \textbf{E}\,=\,essential, \textbf{NR}\,=\,not-relevant. Answer sentences 4 and 5 carry no citations (\texttt{[~]}) because they draw on general clinical knowledge not present in the note.}
\label{tab:example_test2026}
\setlength{\tabcolsep}{4pt}
\renewcommand{\arraystretch}{1.15}
\begin{tabular}{@{}p{0.13\textwidth}p{0.82\textwidth}@{}}
\toprule
\textbf{Field} & \textbf{Case 20 — Neurology}\\
\midrule
\textbf{Patient question} &
\textit{``So what is the dizziness from I never heard of a migraine
that can cause your head to spin like that.''}\\
\addlinespace[2pt]
\textbf{Clinician question} &
\textit{``How did they diagnose her with migraine for spinning sensation?''}\\
\addlinespace[3pt]
\textbf{Clinical note excerpt} &
\textbf{NR}\ \ 1:\ Discharge Instructions: You were evaluated in the Neurology Wards of the \_\_\_ \_\_\_ for headaches that were of an unusual quality.\newline
\textbf{E}\ \ \ \ 2:\ We conducted a series of tests to investigate the cause of your headaches, including a CT scan and an MRI, a series of laboratory tests as well as a lumbar puncture.\newline
\textbf{E}\ \ \ \ 3:\ Through these tests, we were able to rule out serious intracranial causes that can cause a headache, including pseudotumor cerebri, which is a condition where elevated intracranial pressure can cause headaches and subsequent visual damage.\newline
\textbf{E}\ \ \ \ 4:\ We also checked an MRV (MR venogram) which ruled out cerebral venous thrombosis (a blood clot in the veins of your brain).\newline
\textbf{E}\ \ \ \ 5:\ --- We started you on a new medication called verapamil.\newline
\textbf{E}\ \ \ \ 6:\ Take this once daily even if you do not have headaches (everyday, by mouth).\newline
\textbf{NR}\ \ 7:\ --- Inform your PCP of this change, and please follow up with your primary care appointment and neurology appointment.\newline
\textbf{NR}\ \ 8:\ --- You can request your PCP to set up a referral to see a neurologist in the community if you would prefer to do so.\newline
\textbf{E}\ \ \ \ 9:\ --- To achieve optimal control of your migraines, avoid caffeinated products and keep a relatively regular sleep wake cycle (go to bed early, wake up early).\\
\addlinespace[3pt]
\textbf{Gold answer} &
\begin{tabular}[t]{@{}p{0.03\textwidth}p{0.60\textwidth}p{0.14\textwidth}@{}}
\textbf{A1} & Neurology doctors completed a series of tests and procedures to determine the cause of headaches. & cites note \textbf{[2]}\\[3pt]
\textbf{A2} & They ruled out elevated intracranial pressure and blood clots. & cites note \textbf{[3, 4]}\\[3pt]
\textbf{A3} & Verapamil was started once daily to prevent headaches. & cites note \textbf{[5, 6]}\\[3pt]
\textbf{A4} & Based on clinical knowledge, spinning sensation with a migraine is vertigo caused by the migraine. & cites note \textbf{[ ]}\\[3pt]
\textbf{A5} & A migraine with aura can also cause dizziness. & cites note \textbf{[ ]}\\[3pt]
\textbf{A6} & To achieve optimal control of migraines, patient was recommended to avoid caffeine and keep a regular sleep wake cycle. & cites note \textbf{[9]}\\
\end{tabular}\\
\bottomrule
\end{tabular}
\end{table*}

\section{Data}
\label{sec:data}

\subsection{ArchEHR-QA Dataset}

The ArchEHR-QA dataset \citep{soni-demner-fushman-2026-dataset} is derived from de-identified MIMIC-III~\citep{Johnson2016} and MIMIC-IV~\citep{johnson2023mimic} discharge summaries.
Each case comprises four components: a question submitted by a patient
or family member through an online portal; a clinician-interpreted
reformulation for clinical precision; a numbered excerpt from the
relevant note with sentence-level relevance labels (\emph{essential},
\emph{supplementary}, \emph{not-relevant}); and a gold answer of at
most 75~words authored by a clinician.
The dataset is organised into three splits (Table~\ref{tab:dataset}).

The dev split (IDs 1--20) provides clinician answers with full citation annotations and per-sentence relevance labels. For both subtasks, it is used for system fine-tuning. The test-2026 split (IDs 121--167) is the official Subtask~3 evaluation set. Whereas, both the test split (IDs 21--120) and test-2026 split (IDs 121--167) are the official Subtask~4 evaluation set.

Tables~\ref{tab:example_test} and~\ref{tab:example_test2026} present
one dev-split example per subtask, illustrating the full input--output
structure available during system development.
Table~\ref{tab:example_test} (Case~4, Cardiology, MIMIC-III document
180932\_37135) shows the complete 21-sentence note excerpt with gold
relevance labels, used as a Subtask~3 generation example.
Table~\ref{tab:example_test2026} (Case~20, Neurology, MIMIC-IV document
22494097) shows all 9~note sentences with sentence-level relevance labels
and the full citation trail linking each answer sentence to its
supporting note evidence, used as a Subtask~4 alignment example.

%% ===========================================================
\section{System Description}
\label{sec:system}

\subsection{Subtask~3: Answer Generation}
\label{sec:system_t3}

\subsubsection{Base Model}

We use \textbf{Qwen3-4B} \citep{Qwen2025}, a 4-billion parameter
instruction-tuned open-weight language model.
Qwen3's chain-of-thought ``thinking'' mode is disabled via the
\texttt{/no\_think} instruction token, preventing the model from
prepending lengthy reasoning traces that would consume the 75-word
generation budget. The model is loaded in \textbf{4-bit NF4 quantisation} via the
BitsAndBytes library, enabling training on a single consumer GPU.

\subsubsection{QLoRA Configuration}
Full fine-tuning of billion-parameter LLMs requires tens of gigabytes of
GPU memory, prohibitive for most academic settings. LoRA \citep{Hu2022} addresses this by decomposing weight updates into low-rank products $\Delta W = BA$, with $B \in \mathbb{R}^{d \times r}$ and $A \in \mathbb{R}^{r \times k}$ where $r \ll \min(d,k)$, reducing trainable parameters by orders of magnitude while leaving base weights frozen. QLoRA \citep{Dettmers2023} extends this with 4-bit NormalFloat (NF4) quantisation of the frozen base model and 16-bit LoRA adapters, enabling fine-tuning of large-language models on a single GPU with performance comparable to full 16-bit fine-tuning.

LoRA adapters \citep{Hu2022} are attached to all attention and MLP
projection layers:
\texttt{q\_proj}, \texttt{k\_proj}, \texttt{v\_proj}, \texttt{o\_proj},
\texttt{gate\_proj}, \texttt{up\_proj}, and \texttt{down\_proj}.
Table~\ref{tab:hyperparams} lists the full hyperparameter configuration
for both training stages.

\begin{table}[t]
\centering
\small
\caption{Training hyperparameters for Stage~1 (domain grounding) and
Stage~2 (task fine-tuning). LoRA configuration and optimiser are
shared across both stages.}
\label{tab:hyperparams}
\setlength{\tabcolsep}{4pt}
\begin{tabular}{lrr}
\toprule
\textbf{Hyperparameter} & \textbf{Stage~1} & \textbf{Stage~2}\\
\midrule
Base model              & \multicolumn{2}{c}{Qwen3-4B}\\
Quantisation            & \multicolumn{2}{c}{4-bit NF4}\\
LoRA rank $r$           & \multicolumn{2}{c}{32}\\
LoRA $\alpha$           & \multicolumn{2}{c}{64}\\
LoRA dropout            & \multicolumn{2}{c}{0.05}\\
Max sequence length     & \multicolumn{2}{c}{2{,}048}\\
Optimiser               & \multicolumn{2}{c}{AdamW (fused)}\\
LR scheduler            & \multicolumn{2}{c}{Cosine}\\
\midrule
Training samples        & 30{,}000         & 20\\
Epochs                  & 1                & 20\\
Learning rate           & $2{\times}10^{-5}$ & $2{\times}10^{-5}$\\
Effective batch size    & 32               & 4\\
Warmup steps            & 100              & 4\\
\midrule
Initial train loss      & 2.41             & 2.43\\
Final train loss        & 0.149            & 0.860\\
\bottomrule
\end{tabular}
\end{table}

\subsubsection{Two-Stage Training}

\paragraph{Stage~1: Domain grounding.}
The base Qwen3-4B model is fine-tuned for one epoch on 30{,}000
emrQA-MedSQuAD~\citep{Pampari2018} examples.
Training loss falls from 2.41 to 0.149, reflecting rapid acquisition
of clinical language patterns and MIMIC-III note conventions.
The resulting adapter is saved as the initialisation for Stage~2.

\paragraph{Stage~2: Task fine-tuning.}
Beginning from the Stage~1 adapter, the model is fine-tuned for
20~epochs on the 20 annotated development cases.
The small dataset requires careful scheduling: gradient accumulation is
capped to ensure at least five optimiser updates per epoch, and warmup
steps are reduced proportionally.
Final training loss of 0.860 reflects the clinical diversity of the
development cases; we intentionally allow stylistic rather than factual
over-fitting to the target output format.

\subsubsection{Prompt Design}
\label{sec:prompt}

Figure~\ref{fig:prompt} shows the inference prompt template.
Both the patient question and the clinician-interpreted reformulation
are included, allowing the model to target the precise clinical
sub-question while maintaining accessible phrasing.
Note sentences are rendered with their original numeric identifiers
(\texttt{N:~text\textbackslash n}) to preserve the relevance-annotation
indexing used in evaluation.
The system instruction specifies: professional clinical register; a
75-word ceiling; an evidence constraint (\textit{use only information
in the note}); and a prohibition on reproducing structured medication
lists or dosing schedules.

\begin{figure}[t]
\small
\begin{verbatim}
[System]
You are a clinical assistant answering
a patient's question. /no_think
Write a clear, professional answer of
4-5 sentences (max 75 words). Rules:
- Use ONLY information in the note.
- Address what happened and why.
- Do not speculate or add knowledge.
- Do not reproduce medication lists
  or dosing schedules.

Patient Question: <text>
Clinician-Interpreted Question: <text>
Clinical Note Excerpt:
1: Brief Hospital Course:
2: The patient presented with ...
...

Answer:
\end{verbatim}
\vspace{-4pt}
\caption{Subtask~3 inference prompt. The \texttt{/no\_think} token
suppresses Qwen3's chain-of-thought mode. Note sentences are numbered
to preserve relevance-annotation alignment.}
\label{fig:prompt}
\end{figure}

%% -----------------------------------------------------------

\begin{table}[t]
\centering
\small
\caption{Official Subtask~3 test-2026 results (Cases 121--167,
$n$\,=\,47). \emph{Overall} is the arithmetic mean of all six metrics.
\#1 is the top-ranked system; $\Delta$ shows our gap.}
\label{tab:results}
\setlength{\tabcolsep}{4pt}
\begin{tabular}{lrrr}
\toprule
\textbf{Metric} & \textbf{Ours} & \textbf{\#1} & $\boldsymbol{\Delta}$\\
\midrule
BLEU            &  9.42 &  9.92 & $-$0.50\\
ROUGE-L         & 27.04 & 27.85 & $-$0.81\\
SARI            & 55.42 & 58.64 & $-$3.22\\
BERTScore       & 43.00 & 46.83 & $-$3.83\\
AlignScore      & 25.28 & 31.74 & $-$6.46\\
MEDCON          & 37.04 & 43.10 & $-$6.06\\
\midrule
\textbf{Overall}&\textbf{32.87}&\textbf{36.39}&$-$3.52\\
\bottomrule
\end{tabular}
\end{table}

\begin{table*}[t]
\centering
\small
\caption{Official Subtask~4 leaderboard results (Cases 21--120,
147 submissions). \emph{CE alone}: cross-encoder without ensemble.
\emph{Ours}: weighted BM25\,+\,TF-IDF\,+\,CE ensemble with
$\tau_{\text{ens}}=0.85$. \#1 is the top-ranked system;
$\Delta$ shows the gap between our ensemble and \#1.}
\label{tab:t4_results}
\setlength{\tabcolsep}{5pt}
\begin{tabular}{lrrrr}
\toprule
\textbf{Metric} & \textbf{CE alone} & \textbf{Ours (ensemble)} & \textbf{\#1} & $\boldsymbol{\Delta}$\\
\midrule
Micro-precision & 85.46 & 67.35 & 88.0  & $-$20.65\\
Micro-recall    & 47.21 & 66.97 & 75.9  & $-$8.93\\
\textbf{Micro-F1}      & 60.82 & \textbf{67.16} & \textbf{81.5} & $-$\textbf{14.34}\\
\midrule
Macro-precision & 85.09 & 69.33 & ---   & ---\\
Macro-recall    & 51.00 & 69.71 & ---   & ---\\
Macro-F1        & 60.99 & 67.52 & ---   & ---\\
\midrule
\textbf{Overall score} & 60.82 & \textbf{67.16} & \textbf{81.5} & $-$\textbf{14.34}\\
\bottomrule
\end{tabular}
\end{table*}
\subsection{Subtask~4: Evidence Sentence Alignment}
\label{sec:system_t4}

\subsubsection{Problem Formulation}

Given a numbered clinical note excerpt and a gold answer, Subtask~4
requires identifying which sentences constitute the supporting evidence
for that answer.
We model this as binary sentence-level retrieval: each note sentence
receives a relevance score with respect to the answer and is labelled
\emph{cited} or \emph{not-cited}.
Performance is measured by micro-precision, micro-recall, and micro-F1
over all sentence-answer pairs across the 100-case test set.

\subsubsection{Retrieval Methods}

We implement three complementary methods, each producing a relevance
score $s(i) \in \mathbb{R}$ for sentence $i$ given the answer text.

\paragraph{BM25 with relative thresholding (bm25\_rel).}
The answer text is used as a query against the note sentences scored
with BM25 \citep{Robertson1994}.
Because raw BM25 scores depend on note length and vocabulary, we
normalise per case: $\hat{s}(i) = s(i) / s_{\max}$.
A sentence is cited if $\hat{s}(i) \geq \tau_{\text{bm25}}$.

\paragraph{TF-IDF cosine similarity (tfidf).}
Answer and note sentences are represented as TF-IDF vectors;
sentences with cosine similarity $\geq \tau_{\text{tfidf}}$ to the
answer vector are cited.

\paragraph{Fine-tuned cross-encoder (CE).}
We fine-tune \texttt{cross-encoder/ms-marco-MiniLM-L-6-v2} on the
20 development cases using the per-sentence relevance labels as
binary supervision.
To handle gold-empty answers (sentences citing no note evidence), we
augment training with null negatives: for each empty-citation answer
sentence, two randomly sampled note sentences are added as hard
negatives (\texttt{null\_neg\_per\_sent}$=2$).
At inference, raw cross-encoder scores are normalised per note
($\hat{s}(i) = s(i) / s_{\max}$) and sentences with
$\hat{s}(i) \geq \tau_{\text{CE}}$ are cited.

\subsubsection{Weighted Ensemble}

Each method casts a weighted vote per sentence.
Let $w_m$ be the weight of method $m$ and $c_m(i) \in \{0,1\}$
whether method $m$ cites sentence $i$.
The total vote is:
\[
  V(i) = \sum_{m} w_m \cdot c_m(i)
\]
Sentence $i$ is included in the final citation set if
$V(i) \geq \tau_{\text{ens}}$.
Weights are set proportional to individual method precision as observed from post-submission leaderboard feedback, with no access to gold labels during system development:
$w_{\text{bm25}} = 0.527$, $w_{\text{tfidf}} = 0.493$,
$w_{\text{CE}} = 0.855$.
The threshold $\tau_{\text{ens}}$ controls the precision/recall
trade-off and is the primary tuning parameter (Section~\ref{sec:t4_results}).

\paragraph{Method thresholds.}
Individual thresholds $\tau_{\text{bm25}} = 0.50$,
$\tau_{\text{tfidf}} = 0.20$, and $\tau_{\text{CE}} = 0.10$
are set to be permissive---each method casts generous votes---while
the ensemble threshold $\tau_{\text{ens}}$ applies the primary filter.
This separation ensures that CE, despite memorising dev training cases,
can still form high-precision CE+lexical citation pairs for unseen test
sentences.

%% ===========================================================
\section{Experiments}
\label{sec:experiments}

\subsection{Subtask~3: Evaluation Results}

Table~\ref{tab:results} reports our system's performance on the
official test-2026 split (47 cases, IDs 121--167) alongside the
top-ranked system. Our system is competitive on surface n-gram metrics: BLEU within 0.50
points and ROUGE-L within 0.81 points of the top system.
The largest gaps appear on AlignScore ($-$6.46) and MEDCON ($-$6.06),
suggesting our answers, while lexically similar to gold references,
are less tightly grounded to the source note and use less precise
clinical terminology.

%% -----------------------------------------------------------
\subsection{Subtask~4: Evidence Alignment Results}
\label{sec:t4_results}

Table~\ref{tab:t4_results} reports official leaderboard scores for our
submitted ensemble ($\tau_{\text{bm25}}=0.50$, $\tau_{\text{tfidf}}=0.20$,
$\tau_{\text{CE}}=0.10$, $\tau_{\text{ens}}=0.85$) on the 100-case
test set (Cases 21--120, 147 filtered submissions).

The CE-alone baseline is high-precision but low-recall
(P\,=\,85.46, R\,=\,47.21, F1\,=\,60.82): it identifies cited sentences
with confidence but misses the majority.
The ensemble recovers recall at the cost of some precision
(P\,=\,67.35, R\,=\,66.97), yielding a 6.3-point micro-F1 gain
over CE alone (60.82\,$\to$\,67.16).
Macro-F1 (67.52) is marginally higher than micro-F1 (67.16),
indicating that performance is consistent across cases rather
than driven by a small number of high-volume notes.

\section{Conclusion}
\label{sec:conclusion}

We presented a unified system for the ArchEHR-QA Shared Task addressing
both Subtask~3 (answer generation) and Subtask~4 (evidence sentence
alignment).
For Subtask~3, two-stage QLoRA fine-tuning of Qwen3-4B on emrQA-MedSQuAD
and 20 annotated development cases achieves an overall score of
\textbf{32.87} on the test-2026 split.
For Subtask~4, a weighted ensemble of BM25, TF-IDF, and a fine-tuned
cross-encoder with ensemble threshold $\tau_{\text{ens}}=0.85$
achieves micro-F1\,=\,\textbf{67.16}, around 6-point improvement over
the cross-encoder-alone results. The key finding is that adding
bm25+tfidf agreement at a calibrated ensemble threshold recovers
the best precision/recall balance.

Both subtasks point to the same highest-impact future direction. In future work, we plan to scale Stage~1 pre-training beyond the current 30{,}000 emrQA-MedSQuAD~\citep{Pampari2018} examples---the full corpus contains approximately 400{,}000 QA pairs derived from MIMIC-III clinical notes---to further strengthen clinical domain adaptation before task-specific fine-tuning. We also plan to investigate larger cross-encoders to reduce overfitting
in the evidence alignment component. 

\section{References}
\label{sec:reference}
\bibliographystyle{lrec2026-natbib}
\bibliography{12}

\end{document}